\documentclass{article}

\PassOptionsToPackage{numbers, compress}{natbib}

   \usepackage[final]{neurips_2020}

\usepackage{graphics}
\usepackage{epsfig}
\usepackage{verbatim}
\usepackage[utf8]{inputenc} 
\usepackage[T1]{fontenc}    
\usepackage{hyperref}       
\usepackage{url}            
\usepackage{booktabs}       
\usepackage{amsfonts}       
\usepackage{nicefrac}       
\usepackage{microtype}      

\title{Latent neural source recovery via transcoding of simultaneous EEG-fMRI}

\author{Xueqing Liu$^{1}$, Linbi Hong$^{2}$,  and Paul Sajda$^{3}$
\thanks{*This work is supported by the Army Research Laboratory under Cooperative agreement number W911NF-10-2-0022}
\thanks{$^{1}$X. Liu is with the Department of Biomedical Engineering, Columbia University, New York, NY 10027, USA
        {\tt\small xl2556@columbia.edu}}%
\thanks{$^{2}$L. Hong is with the Department of Biomedical Engineering, Columbia University, New York, NY 10027, USA
        {\tt\small lh2517@columbia.edu}}%
\thanks{$^{3}$P. Sajda is with the Departments of Biomedical Engineering, Electrical Engineering and Radiology (Physics), and Member of the Data Science Institute at Columbia University, New York, NY 10027, USA
        {\tt\small psajda@columbia.edu }}%
}

\begin{document}

\maketitle

\begin{abstract}
Simultaneous EEG-fMRI is a multi-modal neuroimaging technique that provides complementary spatial and temporal resolution for inferring a latent source space of neural activity. In this paper we address this inference problem within the framework of  transcoding -- mapping from a specific encoding (modality) to a decoding (the latent source space) and then encoding the latent source space to the other modality. Specifically, we develop a symmetric  method consisting of a cyclic convolutional transcoder that transcodes EEG to fMRI and vice versa. Without any prior knowledge of either the hemodynamic response function or lead field matrix, the method exploits the temporal and spatial relationships between the modalities and latent source spaces to learn these mappings. We show, for real EEG-fMRI data, how well the modalities can be transcoded from one to another as well as the source spaces that are recovered, all on unseen data.  In addition to enabling a new way to symmetrically infer a latent source space, the method can also be seen as low-cost computational neuroimaging -- i.e. generating an 'expensive' fMRI BOLD image from 'low cost' EEG data.  
      
\end{abstract}

\section{Introduction}
Functional magnetic resonance imaging (fMRI) is a neuroimaging modality that is a workhorse for cognitive neuroscience and increasingly used by clinical psychiatric departments \cite{glover2011overview}. fMRI has the advantage of full-brain coverage at relatively high spatial resolution (millimeters), though its temporal resolution is somewhat limited due to the sluggishness of the hemodynamic response \cite{huettel2004functional}. On the other hand, electroencephalography (EEG) is a neuroimaging modality with high temporal resolution (milliseconds) and low spatial resolution as it records electrical signals from electrodes on the surface of the scalp \cite{niedermeyer2005electroencephalography}. In light of such complementarity between the two modalities, when acquired simultaneously, EEG and fMRI potentially can compensate for the shortcoming in one modality using the merits of the other. 

An active area has been the development of machine learning approaches for fusing EEG and fMRI \cite{conroyfusing, oberlin2015symmetrical, transcoding}. One major challenge, however, stems from the fact that EEG and fMRI capture distinct aspects of the underlying neuronal activity, and thus inevitably provide biased and only partially overlapping representations of the latent neural sources  \cite{jorge2014eeg}. An optimal cross modal fusion, should therefore  be capable of 1) identifying the overlapping neuronal substrates for both modalities while minimizing modality-specific bias, and 2) extracting and leveraging any potential information not shared by the two modalities but recorded by either modality, to optimize fusion.

EEG-informed-fMRI 
 \cite{benar2007single, jann2009bold, walz2013simultaneous, muraskin2016brain, muraskin2017fusing, muraskin2018multimodal} 
and fMRI-informed EEG 
 \cite{debener2005trial} 
are the two   approaches for fusing simultaneously acquired EEG-fMRI. They are asymmetrical methods in that only partial information from one modality is used to inform analysis of the other \cite{jorge2014eeg}. EEG-informed fMRI uses features from the EEG to build input regressors in voxel-wise fMRI general linear model (GLM) analyses \cite{jorge2014eeg}. For instance, EEG features can be extracted from trial-to-trial event related potentials \cite{benar2007single}, source dipole time series \cite{muraskin2016brain}, global EEG synchronization in the alpha frequency band \cite{jann2009bold}, or single-trial EEG correlates of task related activity \cite{walz2013simultaneous,muraskin2017fusing,muraskin2018multimodal}. These features are then convolved with a  canonical hemodynamic response function (HRF) before input to a GLM.  The canonical HRF normally peaks around 4 to 5 seconds peristimulus time, lasts 20 to 30 seconds and changes very slowly. It is at best a rough estimate of the hemodynamic coupling with the underlying neural activity, and there is substantial research reporting significant variance in the true HRF, between subjects as well as within a  subject across different brain regions \cite{handwerker2012continuing}.  On the other hand, fMRI-informed EEG applies methods such as fMRI-informed source modeling\cite{debener2005trial} to constrain EEG source localization with a spatial prior provided by information from fMRI. EEG source localization often needs to employ a very complex model of the electromagentic field distribution of the head to calculate the forward and inverse models needed to estimate the location of neural sources in 3D space given the channel recordings on the scalp. These forward and inverse models are based on a leadfield martix which is typically estimated from tissue conductivity and requires complex electromagnetic simulations.

 Symmetrical methods have been developed which treat EEG and fMRI in a more balanced way. For instance, Conroy, et al. \cite{conroyfusing} transformed both EEG and fMRI into the same data space through Canonical Correlation Analysis (CCA). Another example of symmetrical approaches \cite{oberlin2015symmetrical} maps the data fusion problem into an optimization problem, but this approaches still rely on an accurate estimate of the HRF and leadfield matrix, and cannot handle possible non-linearities that may exist between the EEG and fMRI data.

In this paper, we use simultaneously acquired EEG-fMRI and a novel convolutional neural network (CNN) structure to learn the  relationship between EEG and fMRI, and vice versa. Building on a previous work  \cite{transcoding} where a transcoder was developed and tested on simulated data, we substantially develop the approach and leverage the concept of Cycle-Consistent Adversarial Networks (CycleGAN) \cite{zhu2017unpaired}, to create a "cyclic-CNN" for transcoding of simultaneously acquired EEG and fMRI data from an actual EEG-fMRI experiment. The results show that 1) our model can reconstruct fMRI data from EEG data, and vice versa, without any prior knowledge of hemodynamic coupling and leadfield estimates; 2) our model can  accurately estimate the underlying HRF and forward and inverse head models, without prior knowledge of the tissue conductivity or the need for complex electromagnetic simulations. 3) The model can also reveal the dynamics of the latent source space, enabling new ways of assessing the underlying network structure that accounts for the observed EEG and fMRI data.


\section{Methods}
Experimental design, data collection and pre-processing are described in Appendix A.

After pre-processing, our method consists of two steps:
\begin{itemize}
    \item {\it Transcoding the neuroimaging data via a cyclic-CNN:} EEG records the electrical signal from the
surface of the scalp while fMRI records the hemodynamic signals across the whole brain volume.  Since EEG and fMRI are two modalities that record different types of signal, we first decode the signals into a shared latent source space. This is achieved by a concept we term ``neural transcoding`` \-- i.e., generating a signal of one neuroimaging modality from another, by first decoding it into a latent source space and then encoding it into the new measurement space. Transcoding is achieved via a cyclic Convolutional Neural Network (cyclic-CNN as shown in Figure 1). One byproduct of the cyclic-CNN are latent source spaces estimated from EEG and fMRI independently. When transcoding fMRI to EEG, fMRI is first decoded to latent sources, as the fMRI decoder up-sampled the original fMRI in time and transformed it to an electrical signal.  Likewise, when transcoding EEG to fMRI signal, EEG is first decoded to latent sources as EEG decoder spatially up-samples and transforms original EEG  from channel based surface measurements to volumetric data, while the temporal information remains intact. The degree of the up-sampling at this stage, however, is constrained by the computational considerations and data size, this will be discussed further in Session 2.1.  Consequently, the temporal resolution of fMRI was up-sampled only by a factor of 6, with the spatial resolution of EEG up-sampled to approximately 12mm x 12mm x 12mm. 


    \item {\it Upsampling the latent source space via a transformational backprojector:}  We use a transformational backprojector  to combine the transcoded EEG and fMRI data to achieve our integrated source space with EEG's temporal resolution and fMRI's spatial resolution. As the latent source space estimated from fMRI still has higher spatial resolution and lower temporal resolution compared to the latent source space estimated from EEG, we can consider the latent source space estimated from fMRI to be a projection of the integrated latent source space along the time dimension while the the latent source space estimated from EEG as a projection along the spatial dimension. We can thus view the recovery of the integrated latent source space as a sparse, two direction unsupervised back-projection problem. It's "unsupervised" in the sense we do not know the ground truth for the integrated latent source space.  Figure 2 illustrates the transformational backprojector we developed, which uses CNNs and spatial transformer networks (STNs).
\end{itemize}

\subsection{Cyclic-CNN transcoder}

\begin{figure*}[htpb]
\label{pipeline}
  \centering
  
  \includegraphics[scale=0.09]{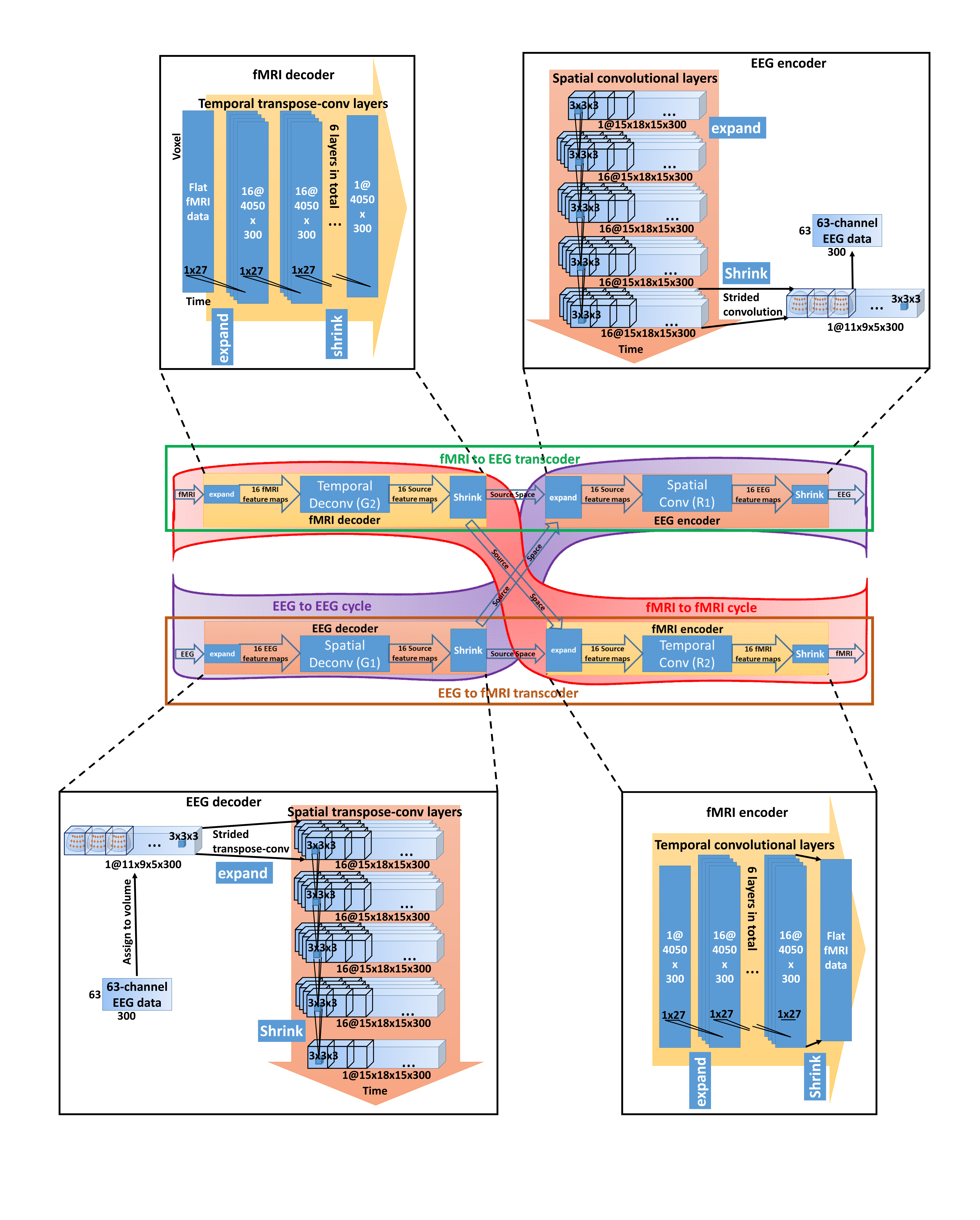}
  \caption{Cyclic convolutional transcoder pipeline}
\end{figure*}
As mentioned, the pre-processed fMRI data was first up-sampled in time by a factor of 6 using linear interpolation, before adjusting for slice timing based on the order of volume acquisition. The fMRI data was then spatially down-sampled by a factor of 6 in each of the x, y, z directions, by taking the mean of each 6 x 6 x 6 block.

The pre-processed EEG channel data is first converted to volumetric data by assigning each channel to a voxel in the volume according to their locations on the EEG cap. Specifically, for our 64 channel EEG cap (where one channel is the ECG channel which is not situated on the scalp), we assign 63 EEG channels to a volume of 11 x 9 x 5 voxels (refer to Appendix E for specific layout of our EEG cap and volume assignment). The EEG data is then temporally down-sampled by a factor of 175, from the sampling rate of 500Hz to 2.86Hz. The degree of such up- and down-sampling is constrained for practical reasons - it allows us to train models and do experiments that are within the computational limits of the GPU RAM and CPU RAM at our disposal.


The fundamental idea of the cyclic-CNN structure is inspired by the cycle-GAN framework proposed by Zhu et al. in \cite{zhu2017unpaired}. A forward mapping from X to the target domain Y, is defined as G: X -> Y, while the reverse mapping from Y to X is defined as R: Y -> X. Cycle consistency loss is defined to enforce $R(G(X)) \approx X$. In our setting, Y is the latent source space, with X being the pre-processed EEG ($E$) or fMRI ($F$) data. Decoding of the EEG and fMRI signals are represented with $G_1$ and $G_2$, respectively; while encoding of EEG and fMRI signals are represented with $R_1$ and $R_2$, respectively. Lastly, $\hat{E}$ represents the estimated EEG through the fMRI-to-EEG transcoder (depicted in the top green rectangular box in Fig.1), $\hat{F}$ represents the estimated fMRI through the EEG-to-fMRI transcoder (depicted in the bottom orange rectangular box in Fig. 1), with $\hat{E'}$ being the estimated EEG through the EEG-to-EEG cycle (depicted in the red shaded cycle in Fig. 1), and $\hat{F'}$ being the estimated fMRI through the fMRI-to-fMRI cycle (depicted in the purple shaded cycle in Fig.1).

The  integrated pipeline is shown in Figure 1. The process of transforming from EEG or fMRI data to the latent source space is a decoding process, and the transforming from the source space to EEG or fMRI data can be considered as an encoding process. The four major modules - the EEG decoder $G_1$, fMRI decoder $G_2$, EEG encoder $R_1$, and fMRI encoder $R_2$ - are all comprised of convolutional layers. Specifically, the encoder modules (EEG and fMRI encoders) are composed of ordinary convolutional layers, while the decoder modules (EEG and fMRI decoders) are composed of transposed convolutional layers which facilitates upsampling in either the temporal or spatial direction, depending on the modality. Among the four modules, the EEG encoder $R_1$ and EEG decoder $G_1$ consist only of spatial convolutional layers, while the fMRI decoder $G_2$ and fMRI encoder $R_2$ consist only of temporal convolutional layers.

With the above specifications, the EEG-to-EEG cycle consistency loss and fMRI-to-fMRI cycle consistency loss can therefore be introduced to enforce $\hat{E'}=R_1(G_1(E)) \approx E$ and $\hat{F'}=R_2(G_2(F)) \approx F$, respectively.  Likewise, the fMRI-to-EEG transcoder consistency loss and EEG-to-fMRI transcoder consistency loss are designed to encourage $\hat{E}=R_1(G_2(F)) \approx E$ and $\hat{F}=R_2(G_1(E)) \approx F$, respectively. The model can thus be represented by four paths with four reconstruction error terms: 

\begin{tabular}{ll}
    Consistency loss of "EEG decoder -> EEG encoder" path: & $loss_1 = \sum_{i=1}^{n}(E_i-\hat{E'_i})^2$ \\
    Consistency loss of "fMRI decoder -> fMRI encoder" path: & $loss_2 = \sum_{i=1}^{n}(F_i-\hat{F'_i})^2$ \\
    Consistency loss of "fMRI decoder -> EEG encoder" path: & $loss_3 = \sum_{i=1}^{n}(E_i-\hat{E_i})^2$ \\
    Consistency loss of "EEG decoder -> fMRI encoder" path: & $loss_4 = \sum_{i=1}^{n}(F_i-\hat{F_i})^2$
\end{tabular}

with the total consistency loss being the sum of the above four losses: $loss_{total}= \sum_{i=1}^4 loss_i$

In the Cyclic-CNN transcoder, the fMRI and EEG spatial and temporal resolutions are adjusted to a computationally tractable resolution for training. This adjustment of resolution does not substantially affect the resolution of data the  transcoder can be applied to as long as the number of channels in EEG and the temporal resolution of fMRI stays the same during the decoding process. Specifically, for fMRI data, the fMRI decoder only applies a temporal transformation to the fMRI data and the same transformation is applied to each voxel. The fMRI decoder therefore can decode testing data of 90 x 108 x 90 voxels (with unit voxel size of 2mm x 2mm x 2mm) the same way as it is used on the training data of 15 x 18 x 15 voxels (with unit voxel size 6 times of that in the testing data, i.e., 12mm x 12mm x 12mm) - as long as the temporal resolution and vector length remains consistent. Similarly, for EEG data, as the EEG decoder only applies a spatial transformation, changing the input data from a sampling rate of 2.86Hz of the training data to 500Hz of the testing data does not affect the EEG decoder's functionality - as long as the number of channels stays the same. With this approach, the original temporal resolution of EEG and spatial resolution of fMRI were both conserved(for more details, refer to Appendix F). While the up-sampled EEG and fMRI signal facilitated feasible computation in the decoding stage, with EEG up-sampled spatially from 63 EEG channels to 15 x 18 x 15 voxels and fMRI up-sampled temporally from TR=2.1s to TR=0.35s (or 2.86Hz).

 
\subsection{Transformational Backprojector}

While a compromise in resolution enables feasible training and testing, we want to recover a high spatio-temporal resolution  latent sources estimates.  The latent source space from fMRI is up-sampled to only 2.86Hz, which is of a substantially lower temporal resolution than the original 500Hz sampling rate of the EEG. Likewise, latent sources estimated from EEG are only expanded to a 3D space of size 15 x 18 x 15 voxels (with unit voxel size of 12mm x 12mm x 12mm), while the original spatial resolution of acquired fMRI is of 2mm x 2mm x 2mm. 

To recover latent source space estimates at the full original resolutions of the neuroimaging data, we fuse the two source spaces through back-projection. Specifically, given a 6 x 6 x 6 x 175 portion of the latent source space, its  field of view is of 12mm x 12mm x 12mm x 0.35s. When projected to the temporal dimension, it will collapse to one time point of size 6 x 6 x 6 x 1. This is equivalent to one volume of the latent source space estimated from fMRI. When projected to the spatial dimensions, the source space will collapse to one time series of length 175, which can be considered as one voxel in the latent source space estimated from EEG. Reversing this process by solving the 2-direction back-projection problem, we combine the latent source space estimated from EEG and the latent source space estimated from fMRI to reconstruct the latent source space with the desired spatial and temporal resolution. The use of this framework is inspired by \cite{backprojector}, where an unsupervised method based on a CNN and spatial transformer network is introduced for solving sparse-view back-projection when there is no access to ground truth. Figure 2 shows an illustration of our framework for latent source space reconstruction with simultaneous EEG-fMRI data.

\begin{figure*}[thpb]
\label{backprojectorpic}
  \centering
  \includegraphics[scale=0.10]{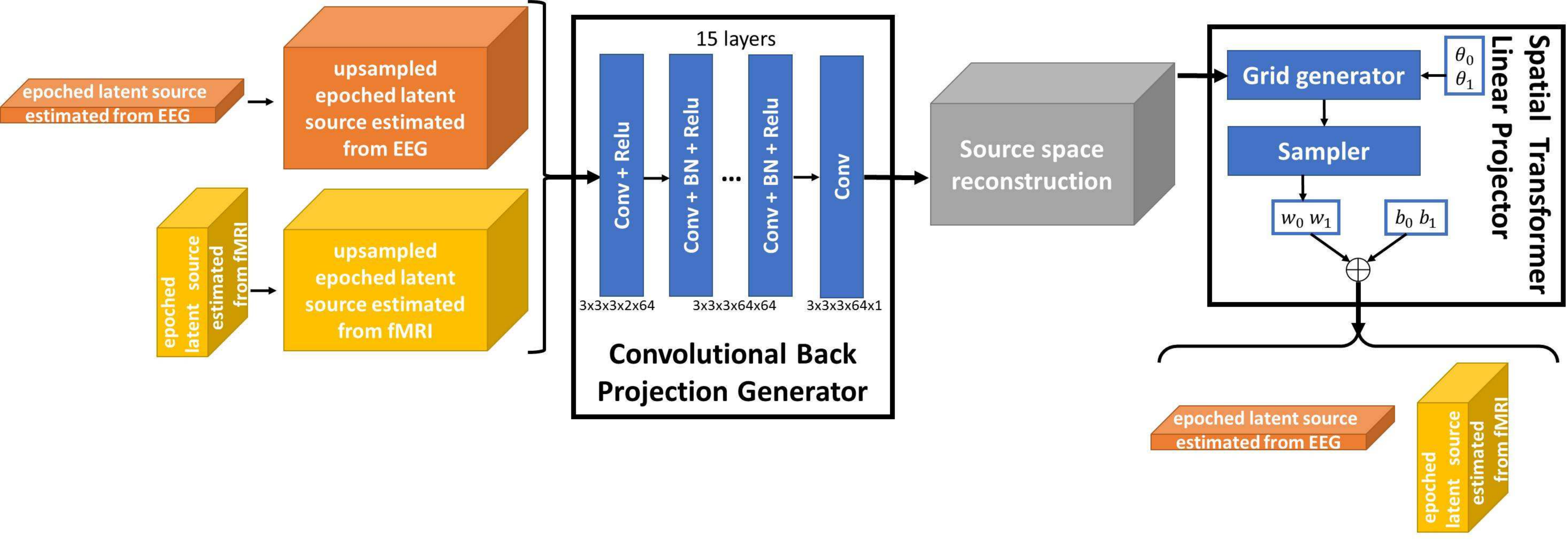}
  \caption{Framework of the Transformational Backprojector}
\end{figure*}

 Furthermore, we apply the backprojector on epoched data (see Experiments and Results). Specifically, the up-sampled epoched source space estimated from EEG and epoched  source space estimated from fMRI serve as two channels of input data, and represent projections from an epoched integrated source space. With each time point considered as one data sample,  every epoch consists of 1200ms (i.e., 600 time points), starting from 350ms before to 850ms after the stimulus. A CNN is used to backproject to the epoched integrated  source space, then a projector is applied to the data to project the estimated integrated  source space back to the source space estimated from fMRI in the direction $\theta_0$ and source space estimated from EEG in direction $\theta_1$.  As temporal backprojection is already achieved by epoching (explained through an illustrative example in Appendix B), in this step, only a spatial transformation is applied. Thus the CNN consists of only spatial convolution layers. The weights $\omega_0, \omega_1$ and biases $b_0, b_1$ are fit to model possible scale and baseline difference between the latent source space estimated from fMRI and the latent source space estimated from EEG.

\section{Training and Testing}
We train and test the transcoder on simultaneous EEG-fMRI data collected from 19 subjects during an auditory oddball experiment \cite{mcintosh2019ballistocardiogram}. We apply a leave-one-subject-out cross-validation, with all sessions of the left-out subject removed during training, and the same procedure repeats for all 19 subjects to obtain group level results. All results reported in the subsequent sections are test results in the left-out subjects. Given the size of the data and computational demands of training, stopping criteria for model training was chosen to be 500 epochs. 

With the approaches outlined above, we carry out two major analyses. One analysis focuses on transcoding the signal from one modality to the other, i.e., to predict fMRI signal from acquired EEG data, and vice versa. The other analysis focuses on identifying the integrated  source space . Note that the resolution of input data varies in these two different types of analysis - specifically, the resolution of EEG or fMRI data is compromised in the transcoding analysis; whereas in the latent source space analysis, resolution of the input data is not adjusted. Specifically, in the first analysis where we test the model's performance on transcoding from one modality to the other, all sessions of all subject's EEG and fMRI data are pre-processed and their spatial and temporal resolution are adjusted to obtain a resolution appropriate for carrying out intensive computation, as described previously. In the second analysis, estimating the integrated latent source space  only involves the readily trained fMRI and EEG decoders, and the transformational backprojector. Thus we  investigate the recovered source space at the full resolutions of the fMRI and EEG. 

In the second analysis, the epoched and up-sampled latent source space estimated from fMRI and latent source space estimated from EEG served as input to the transformational backprojector, while its output is compared with epoched latent source space estimated from fMRI and latent source space estimated from EEG before upsampling. Being an unsupervised method, the transformational backprojector is trained for each session of the epoched data separately, so as to adjust for subject- or session-wise differences of the leadfield. As only a spatial transformation is applied, each volume is treated as one sample in the training. To accelerate the training, the models are initialized with one of the model parameters before training for 50 epochs.

\section{Results}
Here we show the results for the two analyses described above: namely, the performance of transcoding from one modality to the other, and the identification of the latent integrated source space using both EEG and fMRI data. For more details regarding the latent source space estimated from fMRI, see Appendix C. For visualizations of the HRF estimated from data by fMRI encoder module, see Appendix D.

\subsection{fMRI transcoded from EEG}

\begin{figure*}[thpb]
  \centering
  
  \includegraphics[scale=0.18]{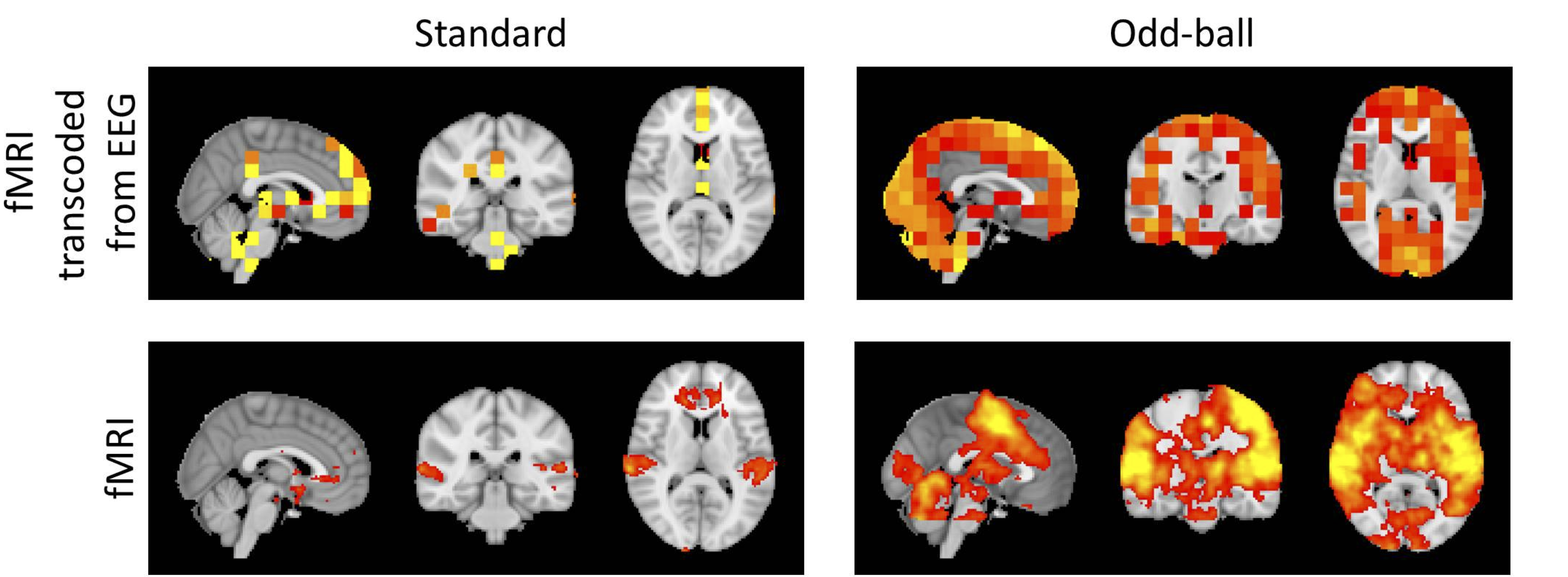}
  \caption{Comparing fMRI transcoded from EEG and real fMRI data: fMRI transcoded from EEG is of spatial resolution of 12mm x 12mm x 12mm while real fMRI data has a spatial resolution of 2mm x 2mm x 2mm}
\end{figure*}

The performance of the EEG-to-fMRI transcoder is shown in Figure 3 using  Z-statistics maps resulted from a group-level  GLM analysis. Specifically we generate fMRI transcoded from the EEG from each left-out subject, and use a GLM analysis with standard and oddball explanatory variables to compare the activation maps between the ground-truth fMRI and the fMRI transcoded from EEG. Figure 3 demonstrates that with fMRI transcoded from EEG, the activation maps show large overlapping voxels (both in terms of the areas of activation and their extent) with that using the actual acquired fMRI data. 

It is commonly believed that EEG source localization is only relatively accurate for cortical regions, since ohmic conduction through the head spatially smears deeper structures\cite{wolters2006influence}. In our analysis, however, the model appears to be able to localize activation in even subcortical regions. Nevertheless, the activation maps using ground-truth fMRI vs. using fMRI transcoded from EEG are not in complete agreement, which is likely due to the fact that there are activities captured by one modality that cannot be detected by the other. In fact, a benefit of our model is that instead of only exploiting the activities in the shared space, as is done in typical asymmetrical EEG-informed fMRI and fMRI-informed EEG analyses, our model attempts to use information even if it is only captured by one of the modalities.

\subsection{EEG transcoded from fMRI}
To evaluate the performance of fMRI-to-EEG transcoder, we calculate the Pearson correlation coefficient between the EEG transcoded from fMRI signal and the real EEG signal (i.e., correlation between $\hat{E}$ and $E$). With respect to the test subjects included in cross-validation, out of a total of 87 runs from 19 subjects, 59 runs show significant correlations at a significant level of alpha=0.05 (significance level determined with Bonferroni correction). Figure 4 shows the fMRI transcoded EEG from a representative channel for one of these runs. Both the ground truth EEG and EEG transcoded from fMRI are shown at the sampling rate of 2.86Hz. The EEG transcoded from the fMRI reasonably matches the ground truth EEG, which is reflective of stimuli induced activity every 2 to 3s. This observation demonstrates that even with the considerably lower  temporal resolution of fMRI (TR=2.1s, i.e., sampling rate of 0.47Hz), the fMRI transcoded EEG signal can still resolve information at higher frequencies. Additionally, the phase and amplitude of the fMRI transcoded EEG signal reflect the estimated forward model of EEG, which are in relatively good agreement with the ground truth. This once  suggests that the transcoder  learns a reasonable  accurate forward model.      
\begin{figure*}[thpb]
  \centering
  
  \includegraphics[scale=0.5]{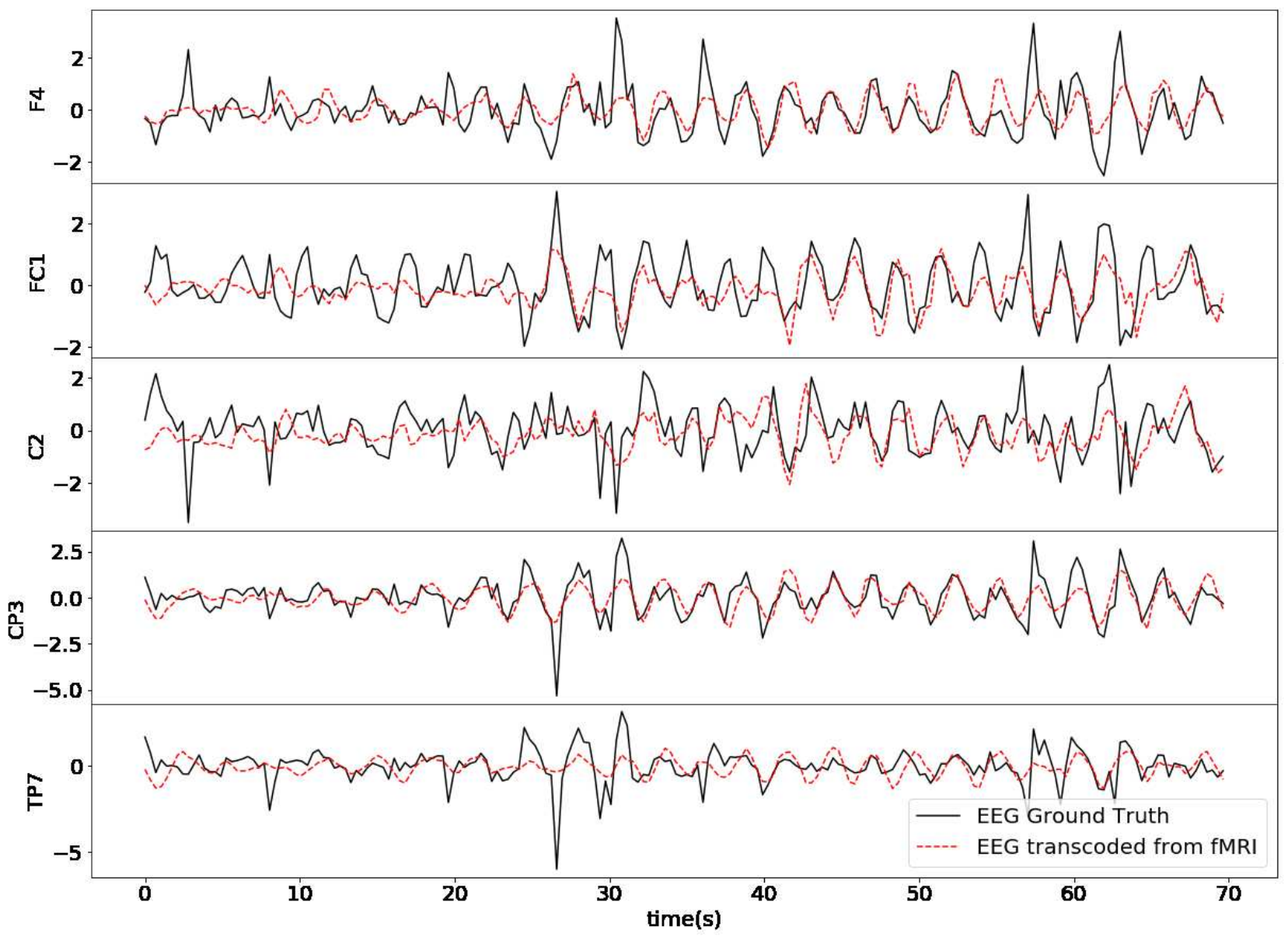}
  \caption{Comparing EEG transcoded from fMRI and real EEG data}
\end{figure*}

\subsection{EEG-fMRI Fused Source Space}
\begin{figure}[thpb]
  \centering
  
  \includegraphics[scale=0.13]{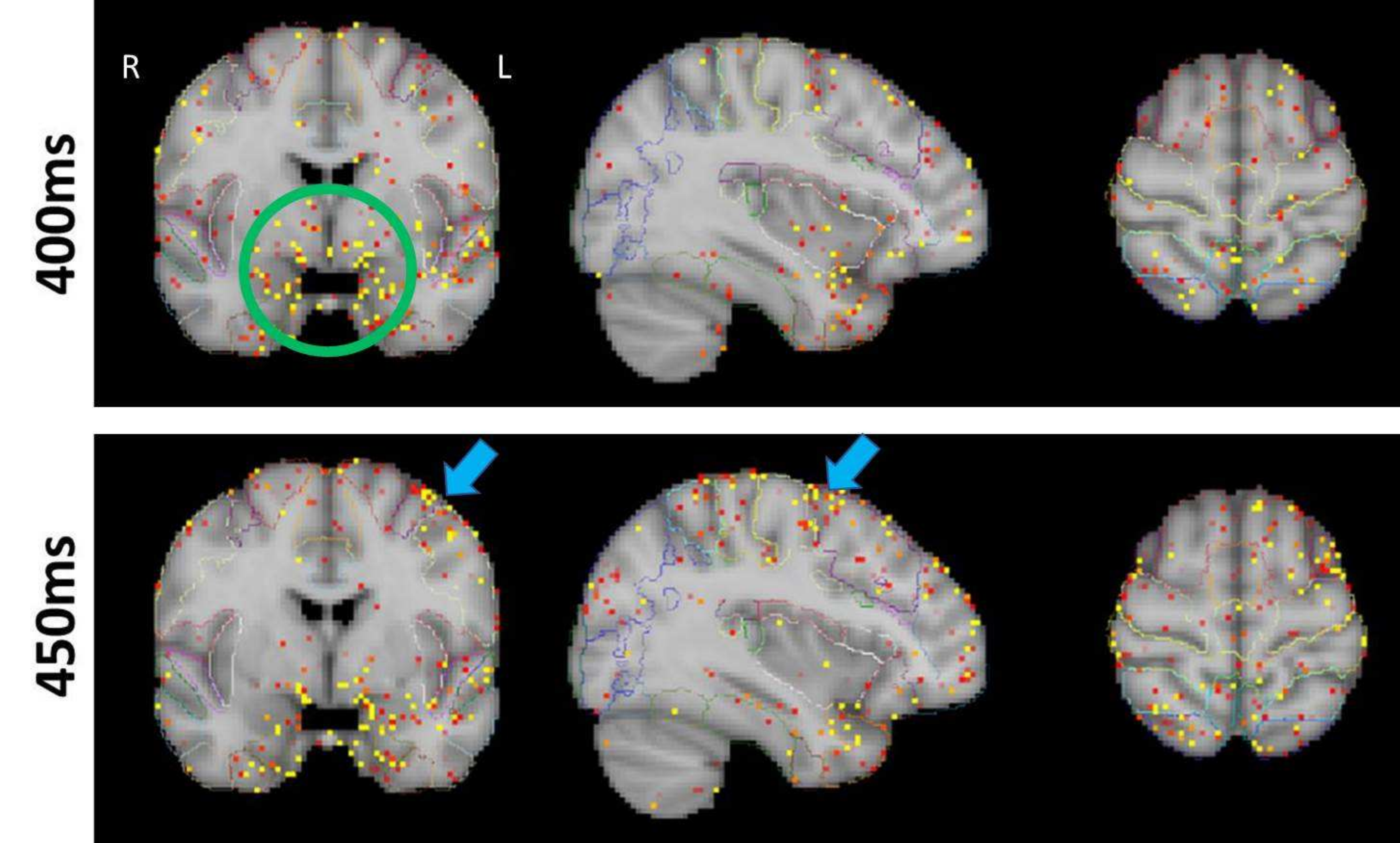}
  \caption{EEG-fMRI integrated source space for oddball stimuli epochs: Shown are sources for two time points leading up to a response (right handed button press) for oddball epochs. Source strength is indicated by color (yellow is stronger than red). The green circle highlights subcortical regions that are part of the motor system, including the putamen, globus pallidus and elements of the basal ganglia associated with voluntary movement. The blue arrows point to activity in the motor cortex that is localized to the right hand (contralateral motor cortex). }
\end{figure}
Figure 5 are representative results of the integrated source  space. The figure shows group-level source activation maps of two volumes at 400ms and 450ms poststimulus. The latent source space has a spatial resolution of 2mm x 2mm x 2mm and temporal resolution of 500Hz. Millisecond scale spatiotemporal brain dynamics associated with the task are included in the Appendix (see video in supplementary material). It is worth noting that the integarted source space exhibits  activation patterns commonly associated with an auditory oddball task. For instance, at approximately 450ms after the onset of the oddball stimulus, activation is seen in primary motor cortex specifically related to finger movement. Preceding this activation, at 400ms is strong  activation in subcortical regions that are part of the motor system. Thus we have evidence that we can resolve cortical and subcortical latent sources at millisecond time scale, that are consistent with the task which cannot be done without integrated information from EEG and fMRI. 

\section{Discussion}
In this paper, we develop a new framework for analysing simultaneously acquired EEG-fMRI data. The framework consists of two parts: A cyclic-CNN that can transcode from EEG to fMRI and vice versa, and a transformational backprojector that can fuse two modalities into a latent source space without compromising the high temporal resolution of EEG or the high spatial resolution of fMRI. More importantly, the models learned after training can be applied to EEG or fMRI that were not acquired at the same time. For instance, if only EEG was acquired in one experimental session, one can apply the trained model to generate the latent source space estimated from EEG, and use such source space to transcode EEG data to the space of fMRI signal. On the other hand, if only the fMRI data are available, one can transcode it to the EEG domain and uncover latent source space estimates from the acquired fMRI signal. 


In contrast to  methods which typically assume linear relationships between latent sources and neuroimaging measures, as well as direct modeling of the hemodynamic responses of the fMRI signal or the electromagnetic characteristics of the human head in advance - our framework is completely data driven and enables non-linear relationships to be learned and modeled. In addition, the model considers simultaneously acquired EEG-fMRI  in terms of a generative problem, and optimizes the model holistically. In our model, the latent source space decoding from EEG or fMRI, the estimation of HRF, and the derivation of leadfield forward and inverse models are all achieved as byproducts during the process of transcoding from one modality to the other.  The fused source space is, in itself, a new neuroimaging data representation, having fMRI's spatial resolution (2mm x 2mm x 2mm) and EEG's temporal resolution (500Hz). This is a resolution never achieved for in vivo non-invasive human brain imaging. This model therefore has the potential of serving as a new tool for human neuroimaging -- i.e. it yields a 3D data volume at a high temporal resolution (see video in Supplementary Material).

Finally, the framework is not simply a blackbox, given in part that we separate spatial and temporal transformations into different modules based on our understanding of the (linear) generative processes. The fMRI decoder, which applies only temporal transpose convolution, can be seen as solving the fMRI deconvolution problem, while the fMRI encoder, constructed of only temporal convolutional layers, can be viewed as the generative process of mapping the latent source to the measurement space via a data-driven estimate of the HRF. Likewise, the EEG encoder, with only spatial transpose convolutional layers, is implementing what other EEG source localization methods normally refer to as the "forward model", while the EEG decoder, having only spatial convolutional layers, is the "inverse model". 

A few caveats and possible improvements over the current work are worth noting. First is that we train a group level model due to the limited data and used leave-one-subject-out to test. However, given more data per subject, we could  potentially train a model for each subject so that subject specific HRF and head electromagnetic characteristics can be learned. Secondly, based on the fact that it is computational expensive to train these models, we only evaluated a small set of hyper-parameters and these have not been optimized. Both of these improvements are part of our future work.


\section{Broader Impact}

The cost of an fMRI scan can range between \$600 to \$1200 and the machine itself, together with necessary support costs, can be well over \$3M.  On the other hand, electroencephalography (EEG) is a neuroimaging modality with high temporal resolution and low spatial resolution which is substantially less expensive, both in terms of scan costs ($<$ \$ 10 per scan) and equipment cost ($\approx$ \$50K). The model we have developed can reconstruct fMRI data from EEG data, and vice versa, without any knowledge of the hemodynamic coupling and leadfield estimates. Possible broader impact is its use as a low cost, computationally-driven approach to produce fMRI images from EEG recordings, thus enabling a \$600 - \$1200 scan to be done at a cost of $<$ \$10. 
\section{Appendix}
\subsection{Appendix A}
\subsubsection{Experimental Design}
We evaluate our cyclic-CNN and transformational backprojector for transcoding EEG-fMRI data using simultaneously acquired EEG-fMRI data from 19 subjects. The data were recorded while subjects performed an auditory oddball task, which included 80\% standard and 20\% oddball
(target) stimuli.  Standard stimuli were pure tones with a
frequency of 350 Hz, while the oddball stimuli were broadband
(laser gun) sounds. Stimuli lasted for 200 ms with an inter-trial
interval (ITI) sampled from a uniform distribution between 2 s
and 3 s. Subjects were instructed to ignore standard
tones and respond to oddball sounds as quickly and as accurately as possible, by pressing a
button. Every subject was scheduled to complete five sessions in total
(105 trials per session), with an average of 4.6 sessions per subject included in this study
(range between 2 to 5, standard deviation of 0.98).

\subsubsection{Data collection and preprocessing}
MR data were recorded inside a 3 T Siemens Prisma scanner, with a 64 channel head/neck coil and EEG was recorded using a 64 channel BrainAmp MR Plus system. Specifically:

\begin{itemize}
    \item Structural T1 images were acquired with an echo time (TE) of 3.95 ms, a repetition time (TR) of 2300 ms, and a flip angle of 9 degrees (FA). Images were acquired with a field of view (FOV) of 176 x 248 voxels, at a voxel size of 1 x 1 x 1 mm. 
    
    \item Functional Echo Planar Imaging (EPI) images were acquired with a TE of 25 ms, a TR of 2100 ms, and an FA of 77 degrees. Images were acquired with a FOV of 64 x 64 voxels, at a voxel size of 3 x 3 x 3 mm.
    
    \item One single-volume high resolution EPI image was acquired with a TE of 30 ms, a TR of 6000 ms, and an FA of 90 degrees. Images were acquired with a FOV of 96 x 96 voxels, at a voxel size of 2 x 2 x 3 mm.
\end{itemize}

For EEG data-processing, raw EEG data was imported with EEGLAB toolbox\cite{delorme2004eeglab} and low-pass filtered with a cutoff frequency of 70 Hz by a non-causal finite impulse response (FIR) filter. An fMRI Artifact Slice Template
Removal algorithm (FASTR) \cite{niazy1999improved} was used for gradient artifact removal. EEG data was then resampled to 500 Hz. To reduce ballistocardiogram (BCG) artifacts, we used the FMRIB EEGLAB plugin. The 500 Hz EEG data was high-pass filtered at 0.25 Hz with another FIR filter to reduce electrode drift before QRS complex detection\cite{christov2004real, kim2004improved} was performed on it. The generated QRS complex event times were used to remove the BCG effect of another copy of the 500 Hz EEG data (not high-pass filtered), with the FMRIB plugin's BCG suppression function set to Optimal Basis Set (OBS) mode with number of bases set to four (the default). This is a widely acceptable standard pipeline for EEG data preprocessing when it's acquired simultaneously with fMRI data \cite{mcintosh2019ballistocardiogram}.

fMRI data was preprocessed using FEAT analysis \cite{woolrich2001temporal} from the FSL toolbox\cite{woolrich2009bayesian}. Specifically, a brain/background threshold was set at 10\% and a high-pass filter with a cutoff at 60s was applied. Spatial smoothing and FSL's built in interleave slice-timing correction were disabled. Spatial normalization was achieved by first registering the fMRI data to a high resolution functional image which is then registered to the structural image (T1) and finally to a standard space image. The fMRI data after spatial normalization is with an FOV of 90 x 108 x 90 voxels, at a voxel size of 2 x 2 x 2 mm.

\subsection{Appendix B: Illustrative example of epoching jittered stimuli}\label{AppB}
\begin{figure*}[htpb]
  \centering
  
  \includegraphics[scale=0.7]{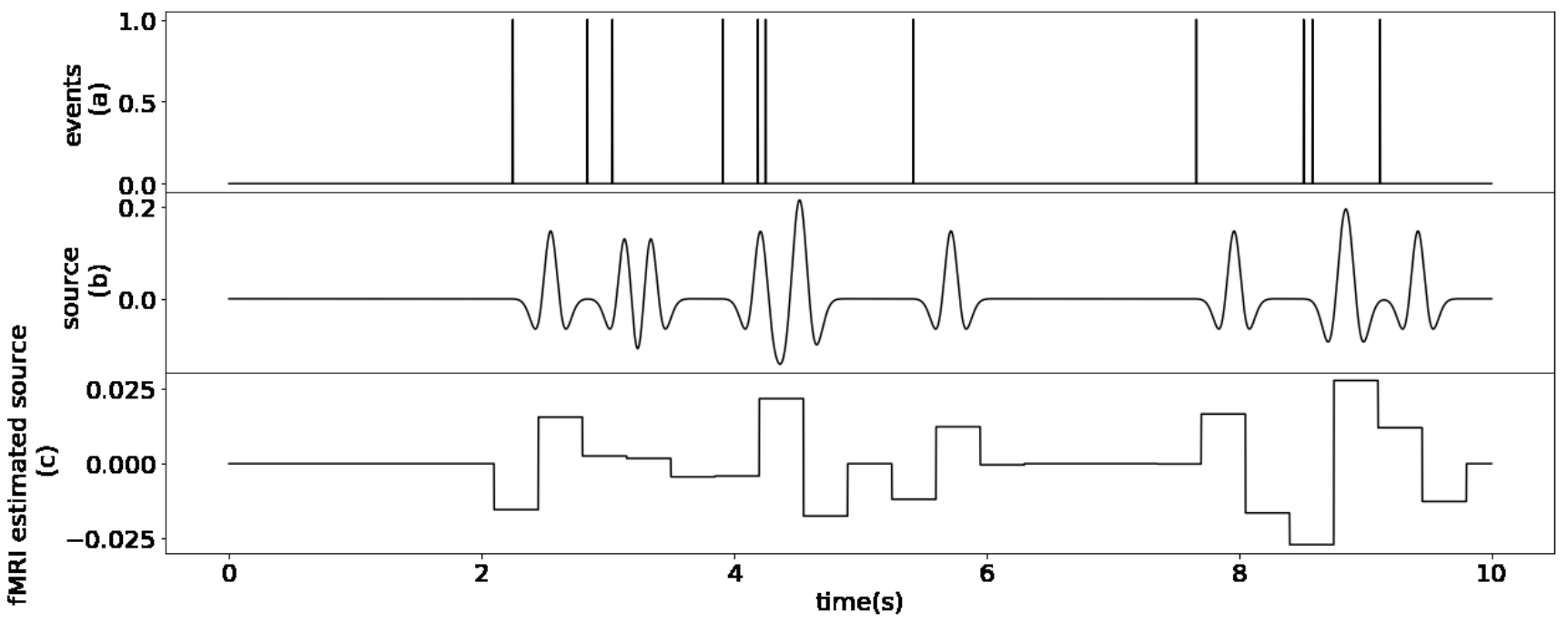}
  \caption{Illustrative example of epoching jittered events: (a) random starting time of stimuli, (b) source time series, (c) fMRI estimated source time series}
\end{figure*}

\begin{figure*}[htpb]
  \centering
  
  \includegraphics[scale=0.7]{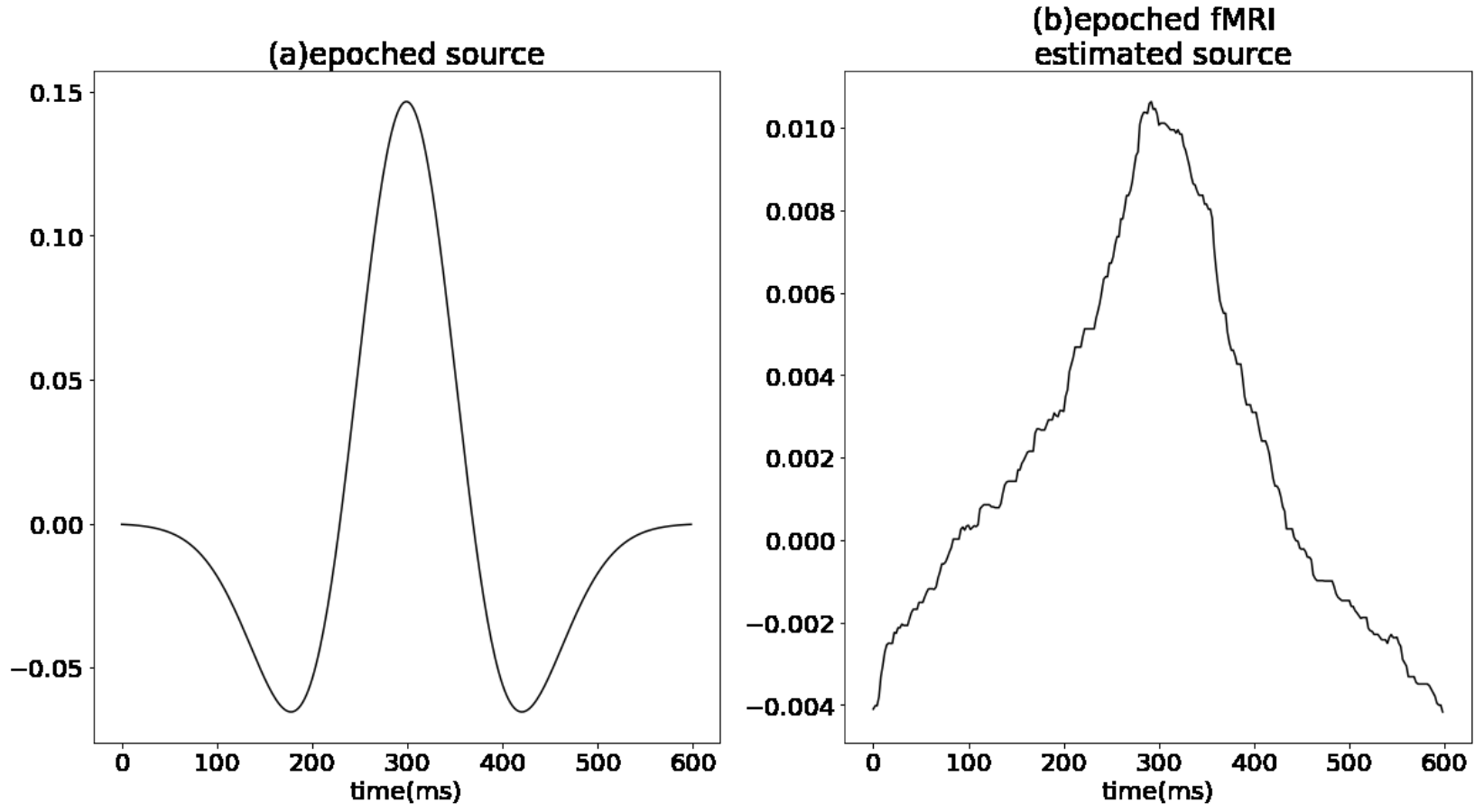}
  \caption{Illustrative example of epoching jittered events: (a) source impulse response, (b) epoched fMRI estimated source}
  
\end{figure*}

Here we show that when stimuli are presented randomly, as is the case in our experiment, epoching the data at a low sampling rate can achieve high temporal resolution and solve the temporal backprojection problem. 
In this illustrative example, we generated 200 random stimuli spanning a time length of 200 s (Figure 1(a) is showing a short period of the simulated stimuli) and convolved these with a  source impulse response as shown in Figure 2(a). The resulting timeseries of the simulated source at 500 Hz is shown  in Figure 1(b). Given the fMRI estimated space has a temporal resolution 2.86 Hz, we can considered it to represent an underlying  500Hz source space projected in the temporal direction for every 175 points. The simulated 2.86 Hz fMRI estimated source signal is thus shown in Figure 1(c), it is difficult to tell what the brain activity looks like as the temporal resolution is too coarse.

Figure 2 shows the original designed source impulse response (a) compared with epoched fMRI estimated source (b) which is an estimate of the source impulse response. Thanks to the jittering nature of the stimuli starting time, the epoched fMRI estimated source signal Figure 2(b) has a finer temporal resolution and resembles the actual source impulse response in Figure 2(a).

\subsection{Appendix C: Latent source space estimated from fMRI}
One of the important features of the framework is the splitting of temporal and spatial convolutional layers into different modules. Applying only a temporal transformation between fMRI and the source space, and applying only a spatial transformation between EEG and the source space makes it possible to interpret the transcoder and decode the latent source space. 

Consider the fMRI decoder as an example. Given that the decoder only applies temporal transpose convolutional layers to the fMRI data, the decoder does not affect the fMRI's spatial information--i.e. it preserves the spatial resolution of the fMRI, while increasing the temporal resolution through the temporal transpose convolutional layers. Therefore, after application of the fMRI decoder, the source space estimation only depends on the properties of the fMRI signal, which we refer to as the fMRI estimated source space. Specifically, the fMRI estimated source space maintains the spatial resolution of fMRI, and has a temporal resolution of 357 ms (1/2.88 Hz) compared to the original resolution of 2.1 s (as TR=2.1 s). The units of fMRI estimated source space are also no longer representative of the  hemodynamic response (as in the original fMRI), rather, it's now representative of an electrical potential (as in EEG).  

Figure 3 shows one of the voxels that resides in the auditory cortex. The original fMRI signal changes slowly, and cannot reflect evoked hemodynamic responses every 2 to 3 seconds as the stimulus happens. The fMRI estimated source, however, is able to capture peaks that align with the auditory input events, and thus provides better temporal resolution for investigating the timing of the evoked neural responses. This result serves as a demonstration that our model can  deconvolve the fMRI data and can achieve better temporal resolution without any modifications to the original EPI data acquisition sequence.

\begin{figure*}[thpb]
  \centering
  
  \includegraphics[scale=0.8]{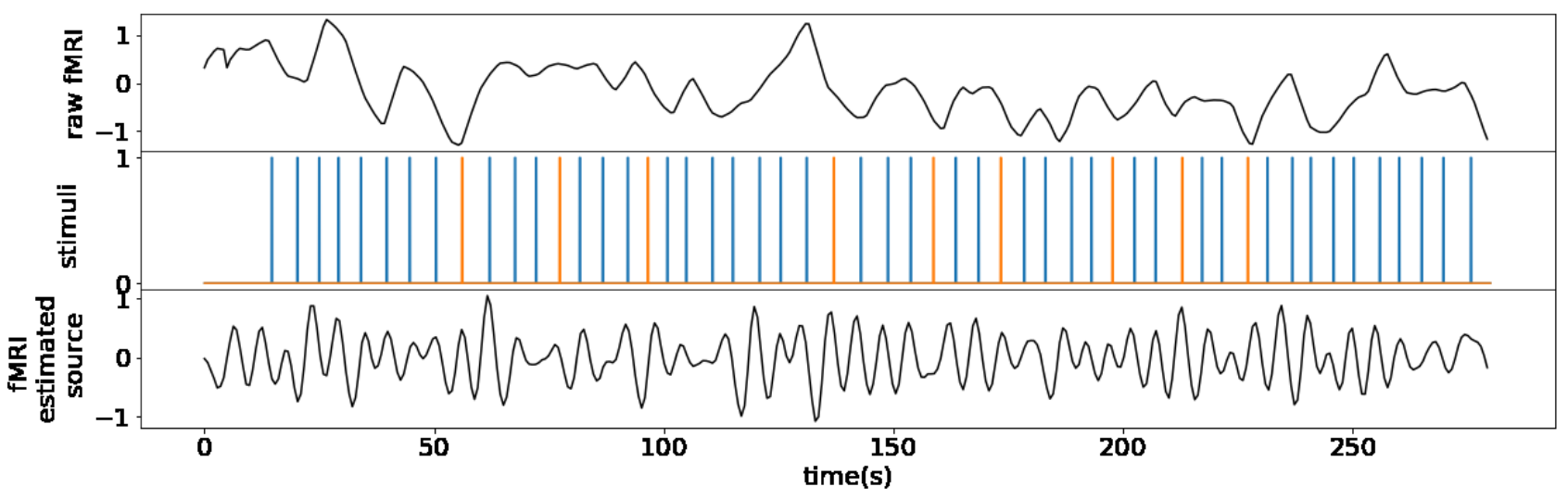}
  \caption{Latent source space estimated from fMRI in auditory cortex}
\end{figure*}

\subsection{Appendix D: HRF estimated by the fMRI encoder module}

\begin{figure*}[thpb]
  \centering
  
  \includegraphics[scale=0.8]{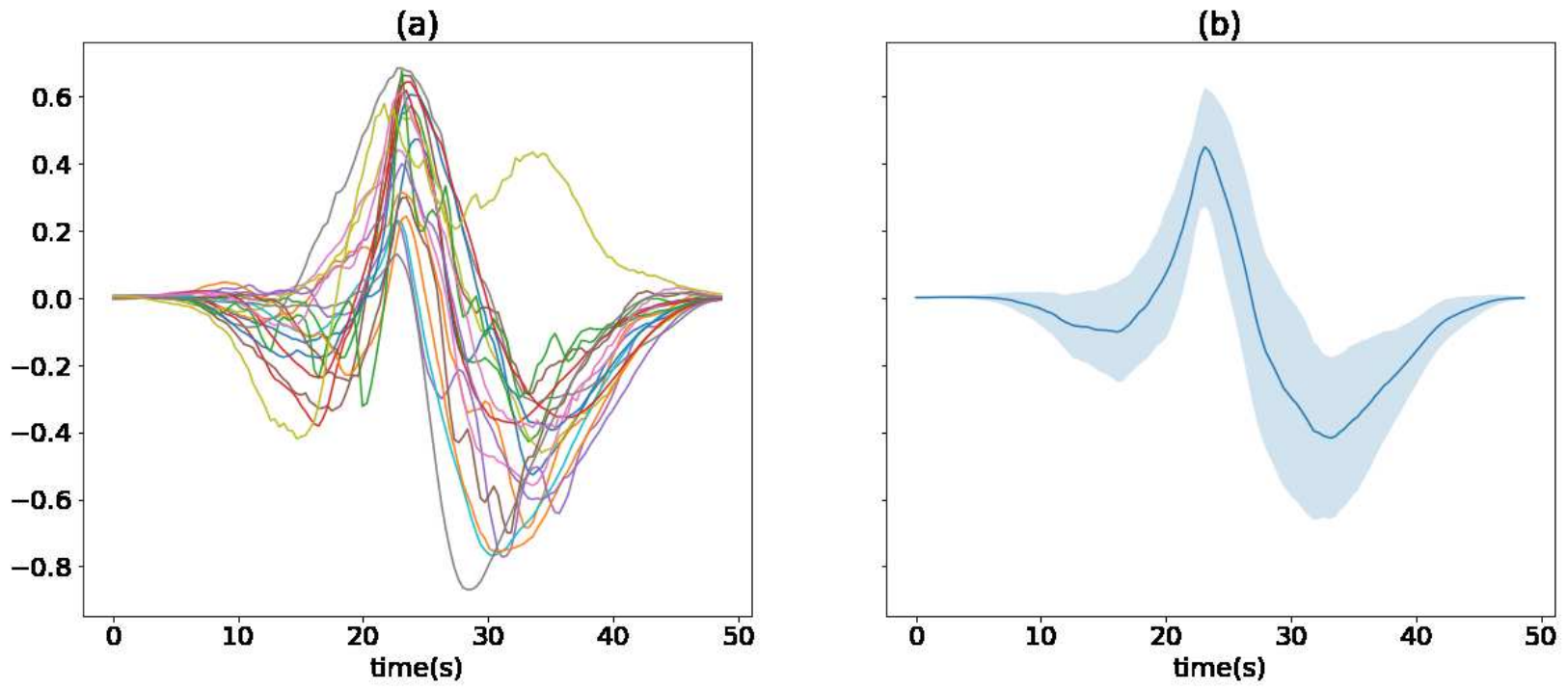}
  \caption{Impulse response function of the fMRI encoder module}
\end{figure*}
The impulse response function learned by the fMRI encoder module is shown in Figure 4.
Since the transformation from fMRI to source space is a deconvolution process, transforming from the source space to fMRI is modeled as the convolution of the source space with a hemodynamics response function (HRF). Unlike traditional general linear model (GLM) analysis methods, however, our method does not define this HRF a priori but instead learns it  directly from the data. Such a data driven framework offers more flexibility than traditional methods and can learn subject-specific HRFs.   

To visualize the HRF estimated by the model, we feed a unit impulse function to the fMRI encoder and observe the output. Figure 4(a) shows the results for all of the 19 models trained using the leave-one-subject-out approach. Figure 4(b) shows the mean HRF across subjects with the 95\% confidence interval. Almost all of the HRFs estimated show clear features of an initial dip, peak and post stimulus undershoot. The timing and timescale of these learned HRFs match the canonical HRF normally used in fMRI analysis, though there is obvious differences across subject that the canonical HRF cannot capture. 

\subsection{Appendix E: Layout of EEG cap and volume assignment}

The layout of our MR compatible EEG cap is shown in Figure 5a. We assigned the 63 electrodes on the surface of the scalp to a 3D volume of size 11 x 9 x 5. This assignment compresses the volume to the smallest meaningful size and ensures consistency in the localization of the electrodes. The outer most ring electrodes in Figure 5a are in the lowest layer of the volume in Figure 5b, and the electrodes in the center ring of Figure 5a are in the top layer. Instead of using a fully connected layer as in \cite{transcoding}, this technique converts channel data to volume data through convolutional layers only and avoids potential overfitting problems. To be noted is that the volume assignment is not required to be very accurate but only need to reflect the rough relative spatial relationship between electrodes. This is because the EEG decoder and EEG encoder of Cycle-CNN transcoder can learn any distortion of the leadfield regardless of the cause of such distortions (be it due to head anatomy or because of this coarse volume assignment).   
\begin{figure*}[thpb]
\label{EEGassign}
  \centering
  
  \includegraphics[scale=0.15]{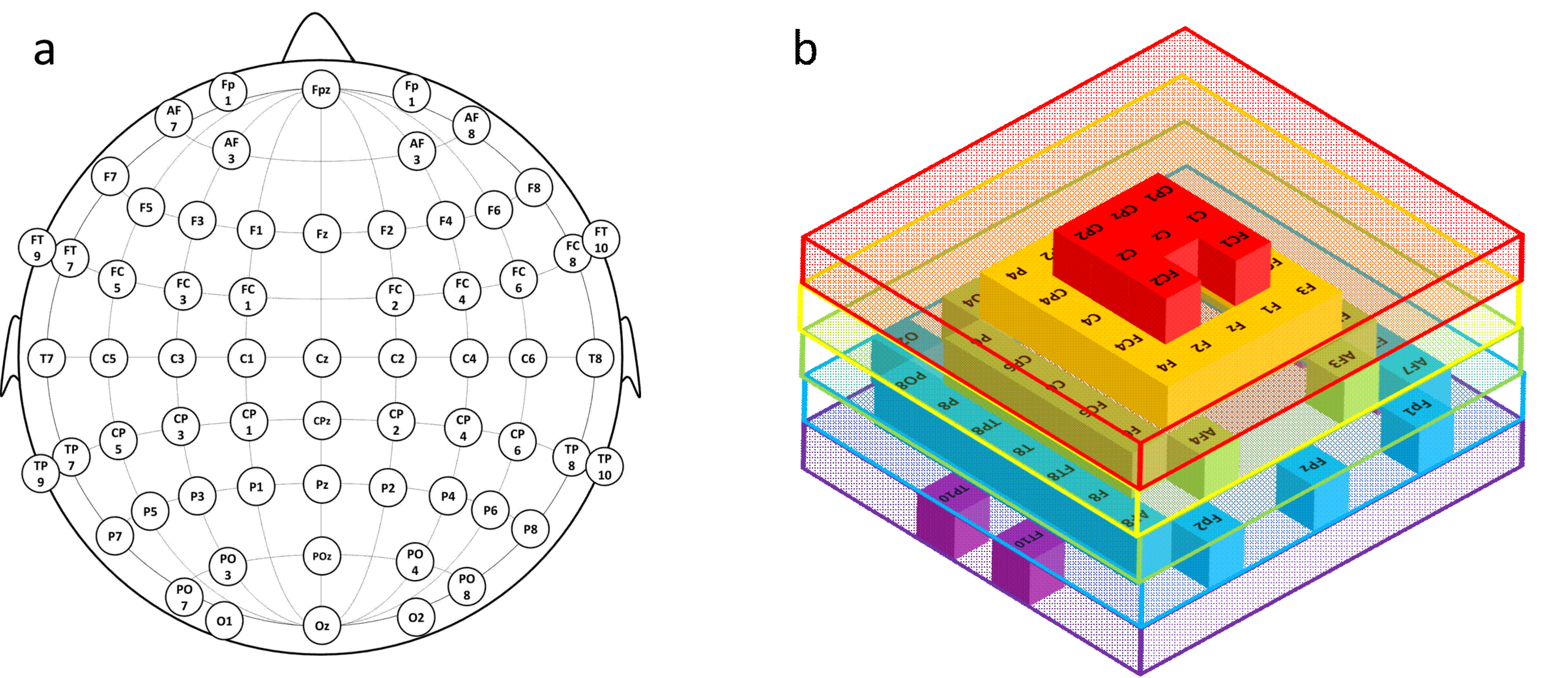}
  \caption{Assignment of  the 63 EEG electrodes channel data to a 11 x 9 x 5 3D volume}
  
\end{figure*}

\subsection{Appendix F: Additional details on training and testing}

The data we used consisted of 87 sessions across 19 subjects. To create an unbiased test set, we trained on data where we left all sessions of one-subject out and then tested on the left out subject's data. This cross-validation was done 19 times so that each subject's data was left out once.

\subsubsection{Cycle-CNN training}

To facilitate training given limitations of the computational resources at our disposal, we created a 'compromise resolution' of our data for training.  Specifically, the original EEG data after preprocessing consists of 63 channels each sampled at 500Hz. The original fMRI data after preprocessing has a spatial resolution of 2 x 2 x 2 mm (volume size 90 x 108 x 90) and TR=2.1 s (0.48 Hz). Since our  goal is to infer an integrated  source space having EEG's temporal resolution (500 Hz) and fMRI's spatial resolution (2 x 2 x 2 mm) this would mean we need to  upsample the original EEG data by a factor of 13886 (90 x 108 x 90 / 63) in volume, and upsample fMRI by a factor of 1050 in time. Though CNNs are known to be good for upsampling/super-resolution, this magnitude of upsampling is challenge for a CNN to learn given relatively limited data. In addition, the size of the data spaces, and network size, make training of such a network computationally prohibitive.

Given these issues we chose a compromise resolution for the EEG and fMRI for training Cycle-CNN transcoder. We downsampled the EEG to 2.86Hz (6 times of original fMRI's sampling rate)  by taking the block-mean of each block of 175 time points, and downsampled the fMRI by taking the block-mean of each 6 x 6 x 6 block of the 90 x 108 x 90 volume to create a volume size of 15 x 18 x 15. The Cycle-CNN transcoder is trained on this resolution of the data. The size of each data sample now is  15 x 18 x 15 x 300 where 300 time points corresponds to 105 s and is long enough to capture the slowly changing hemodynamics. In this setting, the fMRI decoder is only upsampling fMRI by 6 times in the time direction, and the EEG decoder is only mapping from 63 channels (assigned to a volume of 11 x 9 x 5) to a volume of 15 x 18 x 15. At these resolution we can fit the data and model into the memory of the GPUs and CPUs we used. The training code(Cyclic\_CNN\_training.py) is attached in the supplementary material.

\subsubsection{Transformational backprojector training}
The source space estimated from EEG is of spatial resolution of 15 x 18 x 15 (12 x 12 x 12 mm) with a temporal resolution of 500Hz. The source estimated from fMRI is of spatial resolution  90 x 108 x 90 (2 x 2 x 2 mm) and temporal resolution of 2.86Hz. The transformational backprojector is used to fuse these two source space estimations symmetrically for an integrated source space having fMRI's spatial resolution 90 x 108 x 90 (2 x 2 x 2 mm) and EEG's temporal resolution of 500Hz. The transformational backprojector is applied on epoched data, and a new model is trained for each of the 87 sessions. Since the model training is unsupervised this does not introduce any systematic error and also allows the backprojector to capture the inter-subject and inter-session differences. Since we are training for each session, to save training time, we initialize the parameter of each model with one of the trained model's parameter and train each model for 50 epochs. The training code(Transformational\_Backprojector\_training) is attached in the supplementary material. Data and the rest of the code will be uploaded to Github in the future.

\subsection{Appendix G: Video of integrated source space dynamics}
The attached videos (coronal.mp4, sagittal.mp4 and axial.mp4, please play with chrome for the best compatibility.) show the integrated source space dynamics. The three videos correspond to the coronal, sagittal and axial views, respectively. Time 0 of the video is the onset of the stimulus. The sampling rate of the integrated source space is 500 Hz, with each numbered frame corresponds to activity dynamics of 2 ms. The three videos demonstrate a limited yet representative illustration of the integrated source space (as the videos are only showing three slices of the 3D volume). Further interpretation of these results will be discussed in future papers.

\bibliographystyle{IEEEtran}
\bibliography{neurips.bib} 

\end{document}